# Efficient Dense Modules of Asymmetric Convolution for Real-Time Semantic Segmentation

Shao-Yuan Lo[1]   Hsueh-Ming Hang[1]   Sheng-Wei Chan[2]   Jing-Jhih Lin[2]
[1] National Chiao Tung University   [2] Industrial Technology Research Institute
sylo95.eecs02@g2.nctu.edu.tw, hmhang@nctu.edu.tw, {ShengWeiChan, jeromelin}@itri.org.tw

## ABSTRACT

Real-time semantic segmentation plays an important role in practical applications such as self-driving and robots. Most semantic segmentation research focuses on improving estimation accuracy with little consideration on efficiency. Several previous studies that emphasize high-speed inference often fail to produce high-accuracy segmentation results. In this paper, we propose a novel convolutional network named Efficient Dense modules with Asymmetric convolution (EDANet), which employs an asymmetric convolution structure and incorporates dilated convolution and dense connectivity to achieve high efficiency at low computational cost and model size. EDANet is 2.7 times faster than the existing fast segmentation network, ICNet, while it achieves a similar mIoU score without any additional context module, post-processing scheme, and pretrained model. We evaluate EDANet on Cityscapes and CamVid datasets, and compare it with the other state-of-art systems. Our network can run with the high-resolution inputs at the speed of 108 FPS on one GTX 1080Ti.

## 1 INTRODUCTION

Semantic segmentation is an essential area in computer vision. It performs pixel-level label prediction for images. In recent years, the development of deep convolutional neural networks (CNNs) has made notable progress in providing accurate segmentation results [4, 18, 34]. The achievements of these networks mainly rely on their complicated model designs, which consist of considerable depth and width, and need a huge number of parameters and long inference time. However, recent interests in many real-world applications, such as autonomous driving, augmented reality, robotic interaction, and intelligent surveillance, have generated a high demand for the scene understanding systems that are able to operate in real-time. Thus, it is paramount to develop effective convolutional networks for real-time semantic segmentation.

The challenge of designing neural networks by taking both efficiency and reliability into consideration can be seen in Figure 1. For example, most of the top performing methods, such as PSPNet [33] and SegModel [25], focus on improving accuracy at the expense of large increases in computational cost. Therefore, in Figure 1, these methods are located near the area of high accuracy

Project page: https://github.com/shaoyuanlo/EDANet

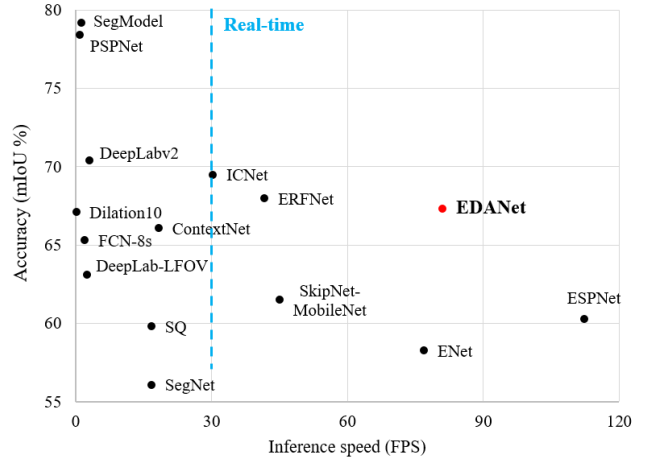

**Figure 1: Inference speed and mIoU accuracy on Cityscapes test set [7]. The speeds are measured on a Titan X. Networks [1,3,4,18,19,21,22,23,25,26,30,32,33,34] are included.**

(mean of intersection-over-union, mIoU) and low inference speed (frames per second, FPS). On the other hand, some approaches, such as ENet [21] and ESPNet [19], emphasize on speed, but their accuracy drops notably. They are located at the bottom right.

In this paper, we propose a new network architecture, Efficient Dense modules with Asymmetric convolution (EDANet), which simultaneously accomplishes high efficiency and accuracy. Our method is not only among the few systems who exceeds 30 FPS (real-time), but also located at the upper right of Figure 1.

One important feature of EDANet is asymmetric convolution. It decomposes a standard 2D convolution into two 1D convolutions. That is, an original $n \times n$ convolution kernel is factorized into two convolution kernels, $n \times 1$ and $1 \times n$, respectively. This technique dramatically reduces the number of parameters with little performance degradation. We take the essence of the densely connected structure [13], and modify it for real-time semantic segmentation. Although DenseNet was initially created for image classification challenges, our experiments show that its capability of gathering the features extracted from different layers and aggregating multi-scale information is innately beneficial to the segmentation task. This structure can also reduce the number of parameters. Dilated convolution is employed by EDANet. The idea is enlarging the receptive fields of networks to retain feature map resolution and avoid losing spatial information. For a good balance in efficiency and reliability, we do not add any extra decoder structure, context module, and post-processing scheme into our system. We further build several EDANet variants to evaluate the performance of different network design choices. In short, EDANet is able to achieve remarkable inference speed and retain high accuracy at the same time.

## 2 RELATED WORK

Originally, CNNs were created for the image classification task [17]. FCN [18] is a pioneering CNN in semantic segmentation. It adapts VGG16 [27] by replacing fully-connected layers with convolution layers to do pixel-level label prediction. Thereafter, semantic segmentation entered the era of CNN-based methods.

**High accuracy networks.** U-Net [24] develops an encoder-decoder architecture to collect spatial information from the shallower layers to enhance the features in the deeper layers. DeconvNet [20] proposes a decoder that is symmetric to its encoder to upsample the outputs of the encoder. These networks have a huge computational cost owing to their heavy decoders. Dilation10 [32] creates a context module by stacking dilated convolution layers with increasing dilation rates for aggregating multi-scale contextual information. DeepLab [4, 5, 6] introduces an atrous spatial pyramid pooling (ASPP) module, which employs multiple parallel filters with different dilation rates to exploit multi-scale representations. Both modules require enormous computation and inference time. As a result, although the aforementioned networks are accurate, they are infeasible for practical applications.

**High inference speed networks.** ENet [21] is one of the first networks aiming at semantic segmentation in real-time. It adapts the ResNet structure [11] but trims the number of convolution filters to reduce computation. ESPNet [19] designs an efficient spatial pyramid (ESP) module, which uses point-wise convolution in front of the spatial pyramids to reduce computational cost. These two networks improve efficiency greatly but significantly sacrifice accuracy. Recent studies, such as ICNet [34] and BiSeNet [31], make a better balance between speed and performance, but there is still room for further improvement.

**Densely connected networks.** DenseNet [13] achieves excellent performance on image classification. Some studies have extended DenseNet to semantic segmentation networks. FC-DenseNet [15] uses DenseNet as an encoder and adds a decoder structure based on the conventional skip connections [24] to build fully convolutional DenseNet. SDN [10] takes DenseNet as their backbone model and combines it with the stacked deconvolutional architecture. These methods simply adopt DenseNet without extensive optimization. Their added complexity further increases the computational cost.

## 3 METHOD

The architecture of the proposed EDANet is shown in Figure 2. It consists of three downsampling blocks, two EDA blocks, and a projection layer. The first and the second EDA block are composed of 5 and 8 densely connected EDA modules respectively. EDANet does not include any additional decoder, context module and post-processing scheme.

In this section, we first describe the core EDA module, then elaborate on the other important network design choices.

### 3.1 EDA Module

The EDA module is the core of the entire EDANet. Its structure is based on the dense module of asymmetric convolution, as shown in Figure 3a. It consists of a point-wise convolution layer and two pairs of asymmetric convolution layers. The output of each EDA module is the concatenation of its input and the newly produced features. Below we discuss each component in the EDA module.

**Point-wise convolution layer.** The point-wise convolution layer is a $1\times1$ convolution at the beginning of each EDA module, which is used to reduce the number of input channels [11]. This design can dramatically decrease the number of parameters and computations.

**Asymmetric convolution.** Asymmetric convolution is to factorize a standard two-dimensional convolution kernel into two one-dimension convolution kernels. In other words, an $n\times 1$ convolution followed by a $1\times n$ convolution can substitute for an $n\times n$ convolution [23, 29]. This mechanism can be expressed as:

$$\sum_{i=-M}^{M}\sum_{j=-N}^{N} W(i,j)I(x-i,y-j) = \sum_{i=-M}^{M} W_x(i)\left[\sum_{j=-N}^{N} W_y(j)I(x-i,y-j)\right] \quad (1)$$

where $I$ is a 2D image, $W$ is a 2D kernel, $W_x$ is a 1D kernel along $x$-dimension, and $W_y$ is a 1D kernel along $y$-dimension. When the kernel size is 3, the number of parameters and computational cost are saved significantly by 33%, and the performance degradation is often very small.

**Dilated convolution.** Dilated convolution is a particular type of convolution, which inserts zeros between two consecutive kernel values along each dimension [3, 33]. This type of convolution can enlarge the effective receptive field of kernels without increasing the number of parameters. For instance, the effective size of an $n\times n$ convolution kernel with dilation rate $r$ is $[r(n-1)+1]\times[r(n-1)+1]$.

In order to aggregate more contextual information for accuracy improvement, we employ dilated convolution at the second asymmetric convolution pair in an EDA module to form the dilated EDA modules, and thus is called "dilated asymmetric convolution". The last two EDA modules in EDA block 1 and all the eight EDA modules in EDA block 2 are the dilated EDA modules. The dilation rates in the system are 2, 2, 2, 2, 4, 4, 8, 8, 16, and 16, respectively. We choose this sequential placement for enlarging the receptive field in a gradual manner.

**Dense connectivity.** Dense connectivity was proposed by DenseNet [13]. We adopt this strategy in EDANet but modify it from layer-level to module-level connections. That is, each EDA module concatenates its input and the new learned features together to form the final output.

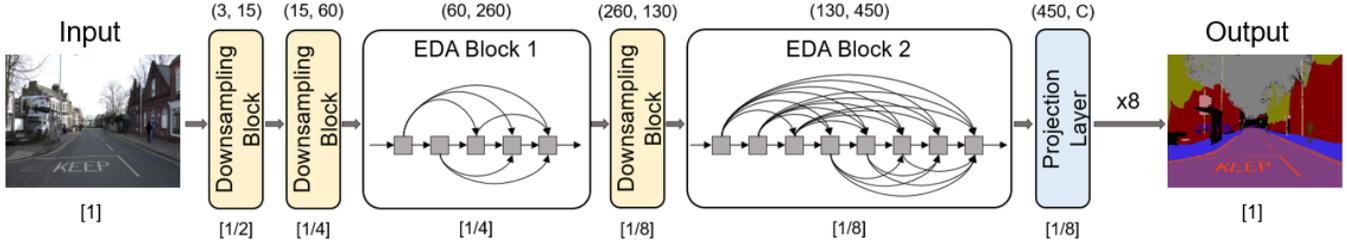

**Figure 2: The proposed EDANet architecture. The numbers of input and output channels of each block are marked in parentheses. The numbers in brackets are output feature size ratios to the full-resolution input images. "C": the number of object classes.**

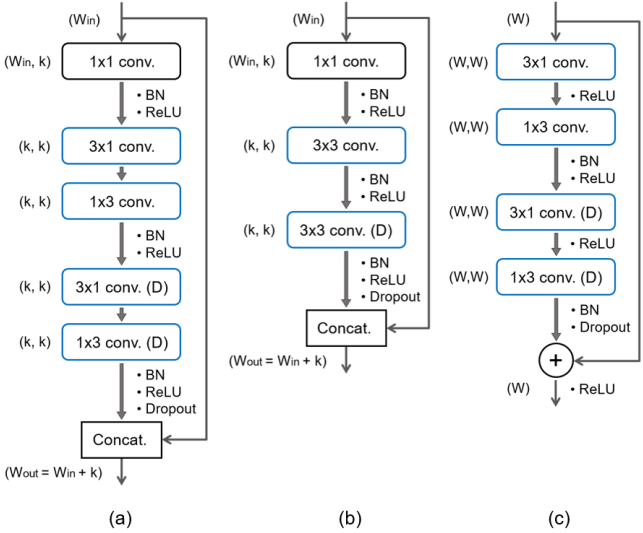

**Figure 3: (a) The proposed EDA module structure. (b) "non-asymmetric" module variant. (c) "non-dense" module variant. The numbers of input and output channels of each layer are marked in parentheses. "(D)": possible dilated convolution. "BN": batch normalization. "k": growth rate, we set it to 40.**

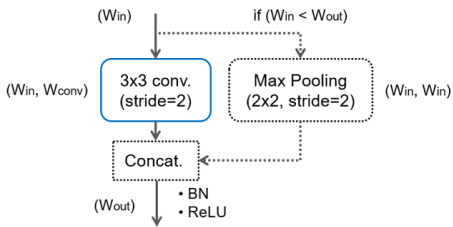

**Figure 4: Downsampling block structure.**

This densely connected structure can substantially increase processing efficiency because each module is only responsible for acquiring a few new features. Furthermore, the deeper layers have larger receptive fields [27]. For example, a stack of two 3×3 convolution layers has the same effective receptive field as a single 5×5 convolution layer, and three such layers have an effective receptive field of 7×7. Thus, the dense connectivity, which concatenate the features learned from each module that has a different receptive field individually, allows our network to naturally gather multi-scale information together. This enables our system to achieve good performance at low computational cost.

### 3.2 Network Design Choices

In this subsection, we discuss the other crucial design choices on the downsampling, decoder, and composite function.

**Downsampling.** We adopt the ENet [21] initial block as our downsampling block. The structure is shown in Figure 4. ENet uses the initial block to do the first downsampling, but we apply it to all the downsampling layers, and further extend it to two modes. When the number of output channels $W_{out}$ of a block is less than the number of input channels $W_{in}$, this block is simply a single 3×3 convolution layer with stride 2, where $W_{conv} = W_{out}$. Our third downsampling block, which divides the network into two EDA blocks, adopts this mode (see Figure 2). If $W_{out} > W_{in}$, a 2×2 max-pooling layer with stride 2 is included, then the concatenation of the features from the convolution and the max-pooling branches forms the final output. In this mode, $W_{conv} = W_{out} - W_{in}$. The first two downsampling blocks adopt this mode (see Figure 2). This two-branch design saves the computation of convolution layers.

The downsampled feature maps enable networks to have larger receptive fields to collect more contextual information. However, reducing feature map resolution would lose spatial details, which is especially harmful to the pixel-wise segmentation. To address this problem, we find a balanced structure, which contains only three downsampling operations in our network. The ratio of the feature size at the end of EDANet to the full-resolution input images is 1/8. Compared to other networks like SegNet [1], whose ratio of feature map size to inputs is 1/32, EDANet remain more spatial details. We use dilated convolution to compensate for the receptive field.

**Decoding.** Many systems use a decoder to upsample feature maps at the expense of huge computation [1, 20]. Even choosing a relatively small decoder still increases the computational cost [23]. Since EDANet aims at fast semantic segmentation, we discard the decoder structure. After EDA block 2, we add a 1×1 convolution layer as a projection layer to output $C$ (the number of classes) feature maps, then use bilinear interpolation to upsample feature maps by a factor of 8 to the size of input images (see Figure 2). This strategy reduces accuracy slightly but saves many computations.

**Composite function.** In order to accelerate the actual inference speed, we choose the traditional post-activation composite function instead of pre-activation [12]. Specifically, the sequence of three operations is a convolution, followed by batch normalization [14] and ReLU. The advantage is that each batch normalization layer can be merged with its preceding convolution layer during inference, which decreases the inference time. In the training phase, we place a dropout layer [28] with dropout rate 0.02 in each module as a regularization measure (see Figure 3).

## 4 EXPERIMENTS

We evaluate our method on two challenging datasets, Cityscapes [7] and CamVid [2]. In this section, we first describe these datasets and our training setup. Then, we conduct a series of experiments to examine the proposed network. Finally, we report the comparisons with the other state-of-art systems.

**Datasets.** The Cityscapes dataset is an urban street scene dataset that contains 19 object classes. It consists 5000 fine-annotated images at the high-resolution of 1024×2048, which are split into three sets: 2975 images for training, 500 images for validation, and 1525 images for testing. There is another set of 19,998 images with coarse annotation, but we only use the fine annotation set for all experiments. Our network is trained and tested on the downsampled 512×1024 inputs. For evaluation, the output features are upsampled by bilinear interpolation to the original dataset resolution.

The CamVid dataset is another dataset for vehicle applications, which consists of 367 training and 233 testing images. It includes 11 classes and has resolution of 360×480.

**Training.** We train our networks by using the Adam optimizer [16] with weight decay 1e$^{-4}$ and batch size 10. We employ the poly learning rate policy, where the learning rate is multiplied by $(1 - iter/\max\_iter)^{power}$ with power 0.9 and initial learning rate 5e$^{-4}$. Inspired by ENet [21], we use the class weighting scheme defined by $w_{class} = 1/log(p_{class} + k)$, where we set $k$ to 1.12. We include data augmentation in training by using random horizontal flip and the translation of 0~2 pixels on both axes. All the reported accuracy results are measured in the mIoU metric.

### 4.1 Ablation Study

In this subsection, we perform a series of experiments to validate the potential of our network. All the following experiments are evaluated on the Cityscapes dataset.

**Core module.** The asymmetric convolution structure and the dense connection concept [13] are two key elements in the proposed EDA module. In order to further investigate potential improvements, we design two variants of our module for comparisons.

The first one is a "non-asymmetric" variant, which replaces the two pairs of asymmetric convolution by two standard 3×3 convolution layers (see Figure 3b). The other one is a "non-dense" variant, which employs the conventional residual connection [11] instead of the dense connection, and it removes the point-wise convolution layer (see Figure 3c). This variant is the same as the ERF module [23]. In order to make comparison at the same computational cost, we set its width $W$ (the number of feature maps) to 40 in block 1 and 80 in block 2 (64 and 128 in ERFNet respectively). We use the same layer placement as EDANet to build two networks composed of the two module variants respectively, and they are called EDA-non-asym and EDA-non-dense.

**Table 1: Ablation study results.**

| Method | mIoU (%) | Params | Multi-Adds |
| --- | --- | --- | --- |
| EDANet | 65.10 | 0.68M | 8.97B |
| (a) Core module. | | | |
| EDA-non-asym | 65.11 | 0.81M | 11.41B |
| EDA-non-dense | 63.92 | 0.73M | 8.87B |
| (b) Extra context module | | | |
| EDA-shallow | 58.09 | 0.55M | 7.77B |
| EDA-ASPP | 60.64 | 3.41M | 41.42B |
| (c) Decoder | | | |
| EDA-ERFdec | 65.56 | 0.78M | 12.95B |
| (d) Downsampling block | | | |
| EDA-DenseDown | 61.63 | 0.42M | 8.51B |

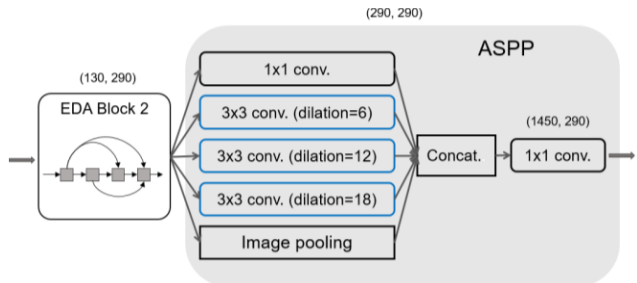

**Figure 5: Part of EDANet-ASPP structure. Image pooling is a global average pooling followed by a 1×1 convolution and bilinear interpolation.**

As shown in Table 1a, EDA-non-asym obtains almost the same accuracy as EDANet, but has 27% more computational cost. This indicates the advantage of our asymmetric convolution design. On the other hand, EDA-non-dense performs 1.18% lower accuracy than EDANet. Apparently, the densely connectivity is effective.

**Extra context module.** Dense connectivity allows EDANet to concatenate multi-scale features and go deeper simultaneously. We compare the ability of our EDA block and the atrous spatial pyramid pooling (ASPP) context module proposed in DeepLab [5] to extract multi-scale representations. We construct EDA-shallow, which contains only four EDA modules in its EDA block 2 as a baseline. Then, we replace the last four EDA modules in EDANet with the ASPP as EDA-ASPP (see Figure 5).

Table 1b shows the results. EDANet attains 7.01% higher accuracy than EDA-shallow, while EDA-ASPP only improves 2.55%. Moreover, EDA-ASPP has 5 times more parameters and 4.6 times more computational cost than EDANet owing to its 5-branch structure. Therefore, we observe that a block of only four connected EDA modules is able to outperform a heavy ASPP context module because of its deeper structure and the excellent capability of aggregating multi-scale information.

**Table 2:** Evaluation results on the Cityscapes test set. The methods whose speed is faster than 15 FPS are included. "†": GTX 1080Ti, with 3,584 CUDA cores and 11,340 GFLOPS. "††": Titan XP, with 3,854 CUDA cores and 12,150 GFLOPS. "†††": Titan X Pascal, with 3,584 CUDA cores and 10,974 GFLOPS.

| Method | Pretrained | mIoU (%) | Speed (FPS) Titan X | Speed (FPS) Other GPUs | Parameters |
|---|---|---|---|---|---|
| SegNet [1] | ImageNet [8] | 56.1 | 16.7 | - | 29.5M |
| ENet [21] | No | 58.3 | 76.9 | - | 0.36M |
| SQ [30] | ImageNet | 59.8 | 16.7 | - | - |
| ESPNet [19] | No | 60.3 | - | 112.9[†††] | 0.36M |
| SkipNet-MobileNet [26] | ImageNet | 61.5 | 45.0 | - | - |
| ContextNet [22] | No | 66.1 | 18.3 | - | 0.85M |
| ERFNet [23] | No | 68.0 | 41.7 | - | 2.1M |
| BiSeNet [31] | ImageNet | 68.4 | - | 105.8[††] | 5.8M |
| ICNet [34] | ImageNet | 69.5 | 30.3 | - | - |
| EDANet (ours) | No | 67.3 | 81.3 | 108.7[†] | 0.68M |

**Table 3:** Evaluation results on the CamVid test set.

| Method | mIoU (%) | Class acc. (%) | Global acc. (%) | Parameters |
|---|---|---|---|---|
| ENet [21] | 51.3 | 68.3 | - | 0.36M |
| ESPNet [19] | 55.6 | 68.3 | - | 0.36M |
| SegNet [1] | 55.6 | 65.2 | 88.5 | 29.5M |
| FCN-8s [18] | 57.0 | - | 88.0 | 134.5M |
| FC-DenseNet56 [15] | 58.9 | - | 88.9 | 1.5M |
| DeepLab-LFOV [3] | 61.6 | - | - | 37.3M |
| Dilation8 [32] | 65.3 | - | 79.0 | 140.8M |
| BiSeNet [31] | 65.6 | - | - | 5.8M |
| ICNet [34] | 67.1 | - | - | - |
| EDANet (ours) | 66.4 | 76.7 | 90.8 | 0.68M |

**Decoder.** After going through the trade-off analysis between efficiency and accuracy, we do not include the decoder structure in our network design. In our investigation, we build a network called EDA-ERFdec, which adds an ERFNet decoder [23], for comparison. This decoder consists of two blocks of a deconvolution layer with stride 2 followed by two ERF modules (see Figure 3c), plus the last deconvolution layer with stride 2 for final output.

As Table 1c shows, EDA-ERFdec obtains 0.46% better accuracy at the expense of 44% more computational cost. Obviously, when we focus on efficiency, adding the decoder does not seem benefit.

**Downsampling block.** We choose the initial block of ENet [21] as the foundation of our downsampling block. Then, we extend it to the two-mode configuration as described earlier. On the other hand, DenseNet [13] uses a 7×7 convolution layer with stride 2 followed by a 3×3 max-pooling with stride 2 for early downsampling, and creates the transition layers that consist of a 1×1 convolution layer followed by a 2×2 average-pooling with stride 2 for the other downsampling operations. For comparing the downsampling approach of ours and the one proposed by DenseNet, we construct EDA-DenseDown by replacing our first two downsampling blocks and the third downsampling block with the early downsampling layers and the transition layer in DenseNet, respectively.

As shown in Table 1d, EDANet attains significantly 3.47% higher accuracy than EDA-DenseDown with only a little more computational cost.

### 4.2 Evaluation Results

We finally train our EDANet in two stages. In the first stage, we train it by the annotations downsampled to 1/8 to the input image size. In the second stage, we train it again by the annotations of the same size as the inputs. In the evaluations, we do not adopt any testing tricks such as multi-crop and multi-scale testing. Table 2 reports our results and the comparisons with the other state-of-art networks in terms of mIoU and inference efficiency on the Cityscapes test set. EDANet achieves 67.3% mIoU, which is better than most of the existing methods that can run at 30 FPS or higher, such as ENet [21] and ESPNet [19], and even outperforms many approaches with lower speed such as Dilation10 [32] and FCN [18]. EDANet attains 108.7 FPS and 81.3 FPS on a single GTX 1080Ti and Titan X, respectively. It is one of the fastest networks now.

We also evaluate our network on the CamVid dataset [2]. As reported in Table 3, EDANet achieves outstanding performance again in efficiency and accuracy. It is able to process a 360×480 CamVid image at the speed of 163 FPS on one GTX 1080Ti card.

## 5 CONCLUSION

In this paper, we have proposed a real-time semantic segmentation network, EDANet, based on the efficient dense modules with asymmetric convolution. The experimental results demonstrate its capability of producing pretty accurate segmentation results with a rather small computational cost comparing to the other state-of-art systems. Going through an extensive investigation, we finally design a well-balanced network architecture for semantic segmentation, which leads to a good trade-off between reliability and efficiency for scene understanding applications.

### ACKOWLEDGMENTS

We would like to thank Ping-Rong Chen for his helpful discussions during the course of this project and Shang-Wei Hung for his drawing for this paper. This work was supported in part by the Mechanical and Mechatronics Systems Research Lab., ITRI, under Grant 3000547822.

# APPENDIX

## A. NETWORK DETAILS

In this appendix, we provide detailed descriptions for the network architectures of the proposed EDANet and all of the variants mentioned in the ablation study section. Tables 4, 5, 6, 7, 8, 9, and 10 correspond to EDANet, EDA-non-asym, EDA-non-dense, EDA-shallow, EDA-ASPP, EDA-ERFdec, and EDA-DenseDown, respectively. In all the following tables, the input sizes are 512×1024. The structures of EDA module, EDA-non-asymmetric module, EDA-non-dense module, downsampling block, and ASPP are shown in Figures 3a, 3b, 3c, 4, and 5, respectively.

Table 4: Layer disposal of the proposed EDANet.

| Name | Mode | Growth rate | # Output channels | Output size |
|---|---|---|---|---|
| Downsampleing block 1 | $W_{in} < W_{out}$ | | 15 | 256×512 |
| Downsampleing block 2 | $W_{in} < W_{out}$ | | 60 | 128×256 |
| EDA module 1-1 | | 40 | 100 | 128×256 |
| EDA module 1-2 | | 40 | 140 | 128×256 |
| EDA module 1-3 | | 40 | 180 | 128×256 |
| EDA module 1-4 | dilation 2 | 40 | 220 | 128×256 |
| EDA module 1-5 | dilation 2 | 40 | 260 | 128×256 |
| Downsampleing block 3 | $W_{in} > W_{out}$ | | 130 | 64×128 |
| EDA module 2-1 | dilation 2 | 40 | 170 | 64×128 |
| EDA module 2-2 | dilation 2 | 40 | 210 | 64×128 |
| EDA module 2-3 | dilation 4 | 40 | 250 | 64×128 |
| EDA module 2-4 | dilation 4 | 40 | 290 | 64×128 |
| EDA module 2-5 | dilation 8 | 40 | 330 | 64×128 |
| EDA module 2-6 | dilation 8 | 40 | 370 | 64×128 |
| EDA module 2-7 | dilation 16 | 40 | 410 | 64×128 |
| EDA module 2-8 | dilation 16 | 40 | 450 | 64×128 |
| Projection layer | 1×1 conv. | | # Classes | 64×128 |
| Bilinear interpolation | ×8 | | # Classes | 512×1024 |
| Bilinear interpolation (inference only) | ×2 | | # Classes | 1024×2048 |

Table 5: Layer disposal of EDA-non-asym.

| Name | Mode | Growth rate | # Output channels | Output size |
|---|---|---|---|---|
| Downsampleing block 1 | $W_{in} < W_{out}$ | | 15 | 256×512 |
| Downsampleing block 2 | $W_{in} < W_{out}$ | | 60 | 128×256 |
| EDA-non-asymmetric module 1-1 | | 40 | 100 | 128×256 |
| EDA-non-asymmetric module 1-2 | | 40 | 140 | 128×256 |
| EDA-non-asymmetric module 1-3 | | 40 | 180 | 128×256 |
| EDA-non-asymmetric module 1-4 | dilation 2 | 40 | 220 | 128×256 |
| EDA-non-asymmetric module 1-5 | dilation 2 | 40 | 260 | 128×256 |
| Downsampleing block 3 | $W_{in} > W_{out}$ | | 130 | 64×128 |
| EDA-non-asymmetric module 2-1 | dilation 2 | 40 | 170 | 64×128 |
| EDA-non-asymmetric module 2-2 | dilation 2 | 40 | 210 | 64×128 |
| EDA-non-asymmetric module 2-3 | dilation 4 | 40 | 250 | 64×128 |
| EDA-non-asymmetric module 2-4 | dilation 4 | 40 | 290 | 64×128 |
| EDA-non-asymmetric module 2-5 | dilation 8 | 40 | 330 | 64×128 |
| EDA-non-asymmetric module 2-6 | dilation 8 | 40 | 370 | 64×128 |
| EDA-non-asymmetric module 2-7 | dilation 16 | 40 | 410 | 64×128 |
| EDA-non-asymmetric module 2-8 | dilation 16 | 40 | 450 | 64×128 |
| Projection layer | 1×1 conv. | | # Classes | 64×128 |
| Bilinear interpolation | ×8 | | # Classes | 512×1024 |
| Bilinear interpolation (inference only) | ×2 | | # Classes | 1024×2048 |

Table 6: Layer disposal of EDA-non-dense. This dilation rate placement is consistent with ERFNet [23].

| Name | Mode | Growth rate | # Output channels | Output size |
|---|---|---|---|---|
| Downsampleing block 1 | $W_{in} < W_{out}$ | | 15 | 256×512 |
| Downsampleing block 2 | $W_{in} < W_{out}$ | | 40 | 128×256 |
| EDA-non-dense module 1-1 | | | 40 | 128×256 |
| EDA-non-dense module 1-2 | | | 40 | 128×256 |
| EDA-non-dense module 1-3 | | | 40 | 128×256 |
| EDA-non-dense module 1-4 | | | 40 | 128×256 |
| EDA-non-dense module 1-5 | | | 40 | 128×256 |
| Downsampleing block 3 | $W_{in} < W_{out}$ | | 80 | 64×128 |
| EDA-non-dense module 2-1 | dilation 2 | | 80 | 64×128 |
| EDA-non-dense module 2-2 | dilation 4 | | 80 | 64×128 |
| EDA-non-dense module 2-3 | dilation 8 | | 80 | 64×128 |
| EDA-non-dense module 2-4 | dilation 16 | | 80 | 64×128 |
| EDA-non-dense module 2-5 | dilation 2 | | 80 | 64×128 |
| EDA-non-dense module 2-6 | dilation 4 | | 80 | 64×128 |
| EDA-non-dense module 2-7 | dilation 8 | | 80 | 64×128 |
| EDA-non-dense module 2-8 | dilation 16 | | 80 | 64×128 |
| Projection layer | 1×1 conv. | | # Classes | 64×128 |
| Bilinear interpolation | ×8 | | # Classes | 512×1024 |
| Bilinear interpolation (inference only) | ×2 | | # Classes | 1024×2048 |

Table 7: Layer disposal of EDA-shallow.

| Name | Mode | Growth rate | # Output channels | Output size |
|---|---|---|---|---|
| Downsampleing block 1 | $W_{in} < W_{out}$ | | 15 | 256×512 |
| Downsampleing block 2 | $W_{in} < W_{out}$ | | 60 | 128×256 |
| EDA module 1-1 | | 40 | 100 | 128×256 |
| EDA module 1-2 | | 40 | 140 | 128×256 |
| EDA module 1-3 | | 40 | 180 | 128×256 |
| EDA module 1-4 | dilation 2 | 40 | 220 | 128×256 |
| EDA module 1-5 | dilation 2 | 40 | 260 | 128×256 |
| Downsampleing block 3 | $W_{in} > W_{out}$ | | 130 | 64×128 |
| EDA module 2-1 | dilation 2 | 40 | 170 | 64×128 |
| EDA module 2-2 | dilation 2 | 40 | 210 | 64×128 |
| EDA module 2-3 | dilation 4 | 40 | 250 | 64×128 |
| EDA module 2-4 | dilation 4 | 40 | 290 | 64×128 |
| Projection layer | 1×1 conv. | | # Classes | 64×128 |
| Bilinear interpolation | ×8 | | # Classes | 512×1024 |
| Bilinear interpolation (inference only) | ×2 | | # Classes | 1024×2048 |

Table 8: Layer disposal of EDA-ASPP. The ASPP structure is consistent with DeepLabv3 [5].

| Name | Mode | Growth rate | # Output channels | Output size |
|---|---|---|---|---|
| Downsampleing block 1 | $W_{in} < W_{out}$ | | 15 | 256×512 |
| Downsampleing block 2 | $W_{in} < W_{out}$ | | 60 | 128×256 |
| EDA module 1-1 | | 40 | 100 | 128×256 |
| EDA module 1-2 | | 40 | 140 | 128×256 |
| EDA module 1-3 | | 40 | 180 | 128×256 |
| EDA module 1-4 | dilation 2 | 40 | 220 | 128×256 |
| EDA module 1-5 | dilation 2 | 40 | 260 | 128×256 |
| Downsampleing block 3 | $W_{in} > W_{out}$ | | 130 | 64×128 |
| EDA module 2-1 | dilation 2 | 40 | 170 | 64×128 |
| EDA module 2-2 | dilation 2 | 40 | 210 | 64×128 |
| EDA module 2-3 | dilation 4 | 40 | 250 | 64×128 |
| EDA module 2-4 | dilation 4 | 40 | 290 | 64×128 |

| ASPP | | | | | | |
|---|---|---|---|---|---|---|
| Branch | (1) | (2) | (3) | (4) | Branch | (5) |
| Convolution | 1×1 | 3×3 | 3×3 | 3×3 | Average-pooling | 64×128 |
| Dilation rate | - | 6 | 12 | 18 | # Output channels | 290 |
| # Output channels | 290 | 290 | 290 | 290 | Output size | 1×1 |
| Output size | 64×128 | 64×128 | 64×128 | 64×128 | Convolution | 1×1 |
| | | | | | # Output channels | 290 |
| | | | | | Output size | 1×1 |
| | | | | | Interpolation | - |
| | | | | | # Output channels | 290 |
| | | | | | Output size | 64×128 |
| Concatenation # Output channels Output size | - 1450 64×128 | | | | | |
| Convolution # Output channels Output size | 1×1 290 64×128 | | | | | |

| Projection layer | 1×1 conv. | | # Classes | 64×128 |
|---|---|---|---|---|
| Bilinear interpolation | ×8 | | # Classes | 512×1024 |
| Bilinear interpolation (inference only) | ×2 | | # Classes | 1024×2048 |

Table 9: Layer disposal of EDA-ERFdec. The decoder structure is consistent with ERFNet.

| Name | Mode | Growth rate | # Output channels | Output size |
|---|---|---|---|---|
| Downsampleing block 1 | $W_{in} < W_{out}$ | | 15 | 256×512 |
| Downsampleing block 2 | $W_{in} < W_{out}$ | | 60 | 128×256 |
| EDA module 1-1 | | 40 | 100 | 128×256 |
| EDA module 1-2 | | 40 | 140 | 128×256 |
| EDA module 1-3 | | 40 | 180 | 128×256 |
| EDA module 1-4 | dilation 2 | 40 | 220 | 128×256 |
| EDA module 1-5 | dilation 2 | 40 | 260 | 128×256 |
| Downsampleing block 3 | $W_{in} > W_{out}$ | | 130 | 64×128 |
| EDA module 2-1 | dilation 2 | 40 | 170 | 64×128 |
| EDA module 2-2 | dilation 2 | 40 | 210 | 64×128 |
| EDA module 2-3 | dilation 4 | 40 | 250 | 64×128 |
| EDA module 2-4 | dilation 4 | 40 | 290 | 64×128 |
| EDA module 2-5 | dilation 8 | 40 | 330 | 64×128 |
| EDA module 2-6 | dilation 8 | 40 | 370 | 64×128 |
| EDA module 2-7 | dilation 16 | 40 | 410 | 64×128 |
| EDA module 2-8 | dilation 16 | 40 | 450 | 64×128 |
| Deconvolution 1 | 2×2, stride 2 | | 64 | 128×256 |
| EDA-non-asymmetric module d1-1 | | | 64 | 128×256 |
| EDA-non-asymmetric module d1-2 | | | 64 | 128×256 |
| Deconvolution 2 | 2×2, stride 2 | | 16 | 256×512 |
| EDA-non-asymmetric module d2-1 | | | 16 | 256×512 |
| EDA-non-asymmetric module d2-2 | | | 16 | 256×512 |
| Deconvolution 2 | 2×2, stride 2 | | # Classes | 512×1024 |
| Bilinear interpolation (inference only) | ×2 | | # Classes | 1024×2048 |

Table 10: Layer disposal of EDA-DenseDown. The dowsampling layers are consistent with DenseNet [13].

| Name | Mode | Growth rate | # Output channels | Output size |
|---|---|---|---|---|
| Convolution | 7×7, stride 2 | | 60 | 256×512 |
| Max-pooling | 3×3, stride 2 | | 60 | 128×256 |
| EDA module 1-1 | | 40 | 100 | 128×256 |
| EDA module 1-2 | | 40 | 140 | 128×256 |
| EDA module 1-3 | | 40 | 180 | 128×256 |
| EDA module 1-4 | dilation 2 | 40 | 220 | 128×256 |
| EDA module 1-5 | dilation 2 | 40 | 260 | 128×256 |
| Convolution | 1×1 | | 130 | 128×256 |
| Average-pooling | 2×2, stride 2 | | 130 | 64×128 |
| EDA module 2-1 | dilation 2 | 40 | 170 | 64×128 |
| EDA module 2-2 | dilation 2 | 40 | 210 | 64×128 |
| EDA module 2-3 | dilation 4 | 40 | 250 | 64×128 |
| EDA module 2-4 | dilation 4 | 40 | 290 | 64×128 |
| EDA module 2-5 | dilation 8 | 40 | 330 | 64×128 |
| EDA module 2-6 | dilation 8 | 40 | 370 | 64×128 |
| EDA module 2-7 | dilation 16 | 40 | 410 | 64×128 |
| EDA module 2-8 | dilation 16 | 40 | 450 | 64×128 |
| Projection layer | 1×1 conv. | | # Classes | 64×128 |
| Bilinear interpolation | ×8 | | # Classes | 512×1024 |
| Bilinear interpolation (inference only) | ×2 | | # Classes | 1024×2048 |

## B. RESULTS ON THE CITYSCAPES AND THE CAMVID DATASETS

In this section, we provide additional segmentation results of the proposed EDANet on Cityscapes [7] and CamVid dataset [2]. First, Tables 11 and 12 list the IoU scores for each class in the two datasets respectively. Then, more visual results are shown in Figures 6 and 7.

Table 11: IoU scores on Cityscapes test set.

| Class | IoU |
|---|---|
| Road | 97.8 |
| Sidewalk | 80.6 |
| Building | 89.5 |
| Wall | 42.0 |
| Fence | 46.0 |
| Pole | 52.3 |
| Traffic light | 59.8 |
| Traffic sign | 65.0 |
| Vegetation | 91.4 |
| Terrain | 68.7 |
| Sky | 93.6 |
| Person | 75.7 |
| Rider | 54.3 |
| Car | 92.4 |
| Truck | 40.9 |
| Bus | 58.7 |
| Train | 56.0 |
| Motorcycle | 50.4 |
| bicycle | 64.0 |
| Metric | Value |
| mIoU classes | 67.3 |
| mIoU categories | 85.8 |

Table 12: IoU scores on CamVid test set.

| Class | IoU |
|---|---|
| Sky | 90.8 |
| Building | 82.5 |
| Pole | 28.5 |
| Road | 93.3 |
| Pavement | 78.3 |
| Tree | 75.0 |
| Sign symbol | 43.7 |
| Fence | 44.4 |
| Vehicle | 81.0 |
| Pedestrian | 54.6 |
| Bike | 57.9 |
| Metric | Value |
| mIoU | 66.4 |
| Class average acc. | 76.7 |
| Global acc. | 90.8 |

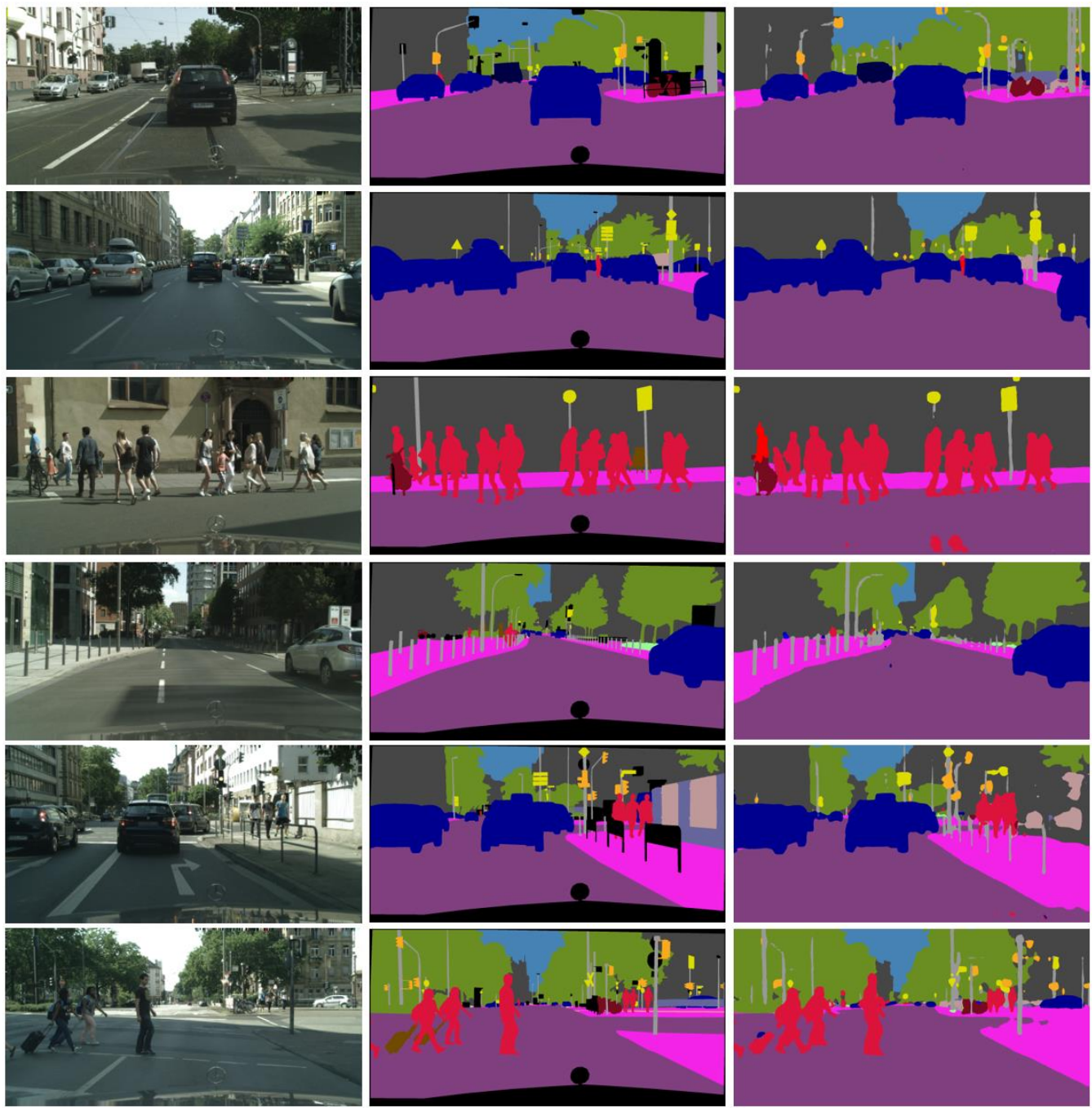

Figure 6: Sample visual results of EDANet on Cityscapes validation set.

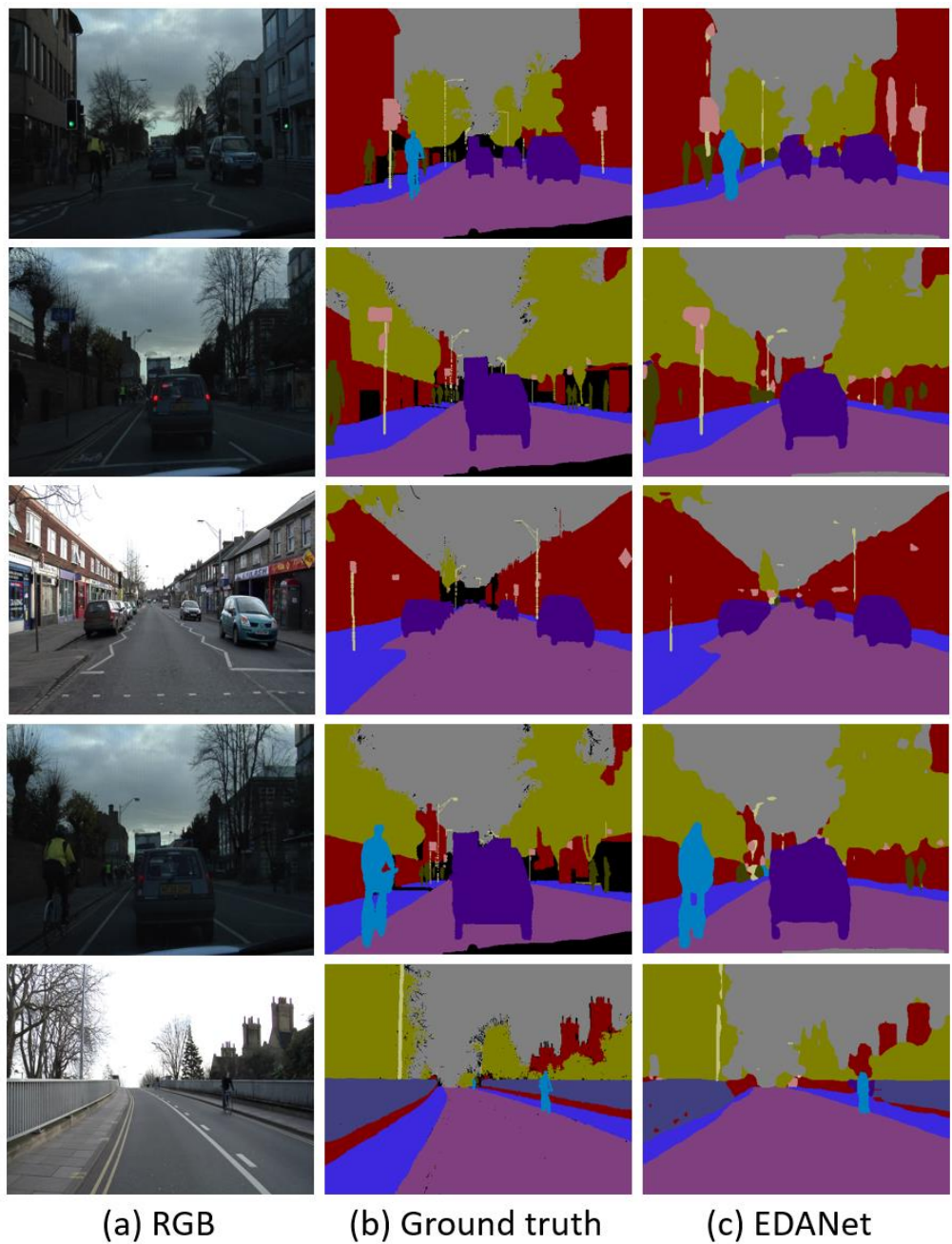

Figure 7: Sample visual results of EDANet on CamVid test set.